# Rapid focus map surveying for whole slide imaging with continues sample motion

Jun Liao,[1] Yutong Jiang,[1] Zichao Bian,[1] Bahareh Mahrou,[1] Aparna Nambiar,[1] Alexander W. Magsam,[1,3] Kaikai Guo,[1] Yong ku Cho,[2] and Guoan Zheng,[1,*]

[1]Department of Biomedical Engineering, University of Connecticut, Storrs, CT, 06269, USA
[2]Department of Chemical & Biomolecular Engineering, University of Connecticut, Storrs, CT, 06269, USA
[3]Department of Biological Systems Engineering, University of Nebraska, Lincoln, NE, 68583, USA

*Corresponding author: guoan.zheng@uconn.edu

**Whole slide imaging (WSI) has recently been cleared for primary diagnosis in the US. A critical challenge of WSI is to perform accurate focusing in high speed. Traditional systems create a focus map prior to scanning. For each focus point on the map, sample needs to be static in the x-y plane and axial scanning is needed to maximize the contrast. Here we report a novel focus map surveying method for WSI. The reported method requires no axial scanning, no additional camera and lens, works for stained and transparent samples, and allows continuous sample motion in the surveying process. It can be used for both brightfield and fluorescence WSI. By using a 20X, 0.75 NA objective lens, we demonstrate a mean focusing error of ~0.08 microns in the static mode and ~0.17 microns in the continuous motion mode. The reported method may provide a turnkey solution for most existing WSI systems for its simplicity, robustness, accuracy, and high-speed. It may also standardize the imaging performance of WSI systems for digital pathology and find other applications in high-content microscopy such as DNA sequencing and time-lapse live-cell imaging.**

With the improvements in digital imaging over the past decade, there has been an upsurge in worldwide attentions on digital pathology using whole slide imaging (WSI) systems, which promises better and faster predication, diagnosis, and prognosis of cancers and other diseases [1]. In particular, the regulatory field for digital pathology using WSI system has advanced significantly in the past years [2]. A major milestone was accomplished early this year when the US Food and Drug Administration approved Philips' WSI system for the primary diagnostic use in the US. The new generation of pathologists trained on WSI systems and the emergence of artificial intelligence in medical diagnosis promises further growth of this field in the coming decades.

Current WSI systems use high-resolution objective lens and mechanical scanning to image different tiles of the sample. The acquired images are then aligned and stitched together to produce a complete and seamless image of the entire slide. The resulting whole-slide image can, thus, provide a quick overview of the entire section, detailed views of areas of interest, and the opportunity to implement machine learning for automatic image analysis. The typical 0.75 NA objective lens used by WSI systems provide the resolution required to resolve structural details. However, their small depth of field poses a challenge to acquire in-focus images of sections with uneven topography. Since different WSI systems use similar objective lens, the autofocusing process is a main influencer of image quality for WSI [3]. Several studies have implicated poor focus as the main culprit for poor image quality in WSI [4, 5].

To address this challenge, current WSI systems create a focus map prior to scanning. For each focus point on the map, a traditional WSI system will scan the sample to different focal planes along the z-axis and acquire a z-stack (as many as 20 images are needed). The z-stack will then be analyzed for a figure of merit, such as image contrast or entropy, to identify the ideal focal point for one tile position. This process will be repeated for other tiles of the whole slide image. Since a typical whole slide image contains more than 400 tiles, surveying the focus points for every tile would require a prohibitive amount of time for high throughput scanning. Most existing systems select a subset of tiles for focus point surveying or skip every 3-5 tiles to save time. The focus points of the selected tiles are then triangulated to re-create the focus map of the entire tissue section.

This well-established focus map surveying method suffers from three challenges. First, the assumption with skipping tiles is that adjacent tiles share the same focal position. However, it has been shown that the focal positions of two adjacent tiles can vary by more than 1 μm [6]. Therefore, this assumption, in fact, is not true. Skipping tiles will lead to a poor focusing accuracy and poor image quality. Yet more focus points come at the expense of decreasing speed. Many WSI systems allow the user to select the number of focus points to create a map. Second, this focus point surveying method relies on maximizing the image contrast of the z-stacks. Many pathology samples, however, are weakly stained and the image contrast is low. Some immunohistochemistry slides are even transparent under brightfield illumination. It is challenging to handle these cases using the current focus map surveying method. Third, the focus point surveying process requires the sample to be static during the acquisition process. In other words, it is challenging to recover the focus point while the sample is in continuous x-y motion. Motion accelerating and deaccelerating would substantially decrease the scanning speed.

Table 1. Summary of different focusing methods in WSI systems

|  | Conventional method | Dual sensors [2, 6] | OCT method [10] | This work |
|---|---|---|---|---|
| Hardware | One camera | Two or more cameras | Complex Fourier-domain OCT setup | One camera with 2 LEDs |
| Principle | Maximize image contrast through axial scanning | Additional cameras for focus point tracking | OCT A-scan to locate the axial position | Convert axial information to lateral image shift |
| Data processing | Calculate a figure of merit | Calculate a figure of merit | FFT to get the axial profile | 1D image autocorrelation |
| Sample | Only works for high-contrast sample | Only works for high-contrast sample | Works for transparent samples | Works for transparent samples |
| Modality | Brightfield only | Brightfield only | Brightfield and fluorescence | Brightfield and fluorescence |
| Focusing z-range | ~10 μm, determined by the number of z-stack images | < 10 μm | > 100 μm, determined by the light source and spectrometer | > 60 μm, determined by the spatial coherence of the LEDs |
| Speed | Slow due to acquiring z-stack | Fast | Fast | Fast |
| Motion | x-y motion blur not allowed | Slow x-y motion allowed with pulsed illumination | x-y motion allowed | x-y motion allowed without pulsed illumination |

Some recent innovations in WSI systems are able to tackle the challenges listed above. For example, the dual sensor method is able to perform dynamic focusing while the sample is in continuous motion [3, 6]. In this approach, the light from the sample is split to two cameras. One is for capturing the high-resolution image of the sample and the other is rapidly scanned through 3 different planes to locate the best focal plane position. This approach requires fast axial scanning and cannot handle transparent samples. Our group has also demonstrated the use of one or two additional cameras and additional lenses to perform dynamic autofocusing [7, 8]. The use of the additional camera system and its alignment to the microscope may not be compatible with most existing WSI platforms. Another interesting approach from a recent WSI system (Thorlabs EnVista) is to use optical coherent tomography (OCT) to get an A-scan of the sample's axial profile [9]. The focal position can then be identified from the A-scan. This approach is able to handle transparent samples. However, complicated Fourier-domain OCT hardware is needed. We summarize the key considerations in Table 1.

In this letter, we report a novel focus map surveying method for WSI. In this method, we illuminate the sample with two incident angles and recover the focus points for every tile without axial sample scanning. To survey the focus points under continuous sample motion, we explore the unique 1D autocorrelation strategy of the reported method. By choosing the scanning direction to be perpendicular to the autocorrelation direction, we can minimize the effect of motion blur. We have tested the reported approach on 600 tiles on 10 pathology samples, including transparent and low-contrast samples. We demonstrate a mean focusing error of ~0.08 microns in the static mode and ~0.17 microns in the continuous motion mode. The reported method requires no axial scanning, no additional camera and lens, works for stained and transparent samples, and allows continuous sample motion in the surveying process. It may provide a turnkey solution for most existing WSI systems for its simplicity, robustness, accuracy, and high-speed. It may also standardize the imaging performance of WSI systems for digital pathology and find other applications in high-content microscopy such as DNA sequencing and time-lapse live-cell imaging.

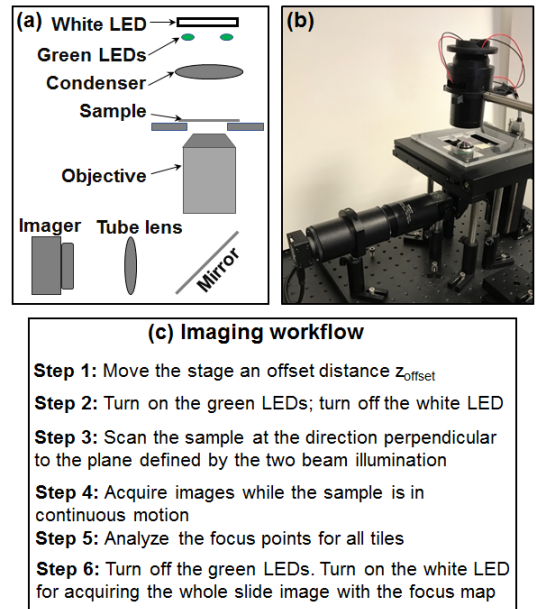

Fig. 1. (a) The scheme of the proposed WSI platform (Visualization 1). (b) The experimental prototype setup. (c) The workflow.

Figure 1(a) shows the reported focus map surveying scheme. The core components are the same as a regular microscope. We only need one camera for both surveying the focus map and acquiring high-resolution images. At the illumination path, two LED elements (Luxeon Rebel Green LEDs) are placed at the back focal plane of the condenser lens and they illuminate the sample with two oblique incident angles. The LED elements can be treated as partially coherent light sources and they are able to generate coherent contrast for samples at out-of-focus positions.

The workflow of our method is shown in Fig. 1(c). In step 1, we move the sample to a pre-defined offset position $z_{offset}$. This step serves two purposes. First, it generates out-of-focus contrast using the partially coherent LED illumination. Second, it facilitates the autocorrelation analysis of the focus point (if $z_{offset}$ is too small, the autocorrelation peaks cannot be accurately located). In our experiment, we choose an offset position of 60 μm. In step 2, we turn off the white surface mounted LED and turn on the two green LED elements. If the sample is placed at a defocus position, the captured image from the main camera will contain two copies of the sample separated by a certain distance. By identifying this distance, we can recover the focus plane of the sample [7, 8]. In step 3, we scan the sample in the direction that is perpendicular to the plane defined by the two LED illumination. In this case, the scanning direction is perpendicular to the two-copy direction of the captured image. The motion blur effect has a minimum impact on the recovered focal position (also refer to Fig. 3). In step 4, we acquire images while the sample is in continuous motion. These images will be analyzed to generate the focus map of the sample (Visualization 1). In step 5, we move the sample to the correct positions based on the focus map and acquire the high-resolution whole-slide image.

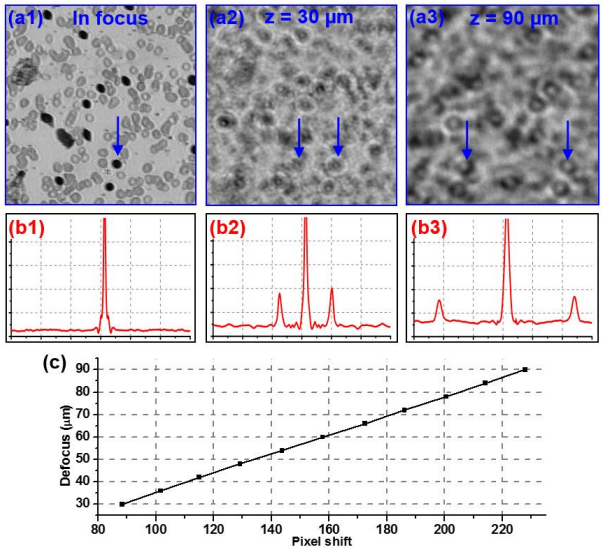

Fig. 2. (a) Images captured with two LED illumination. (b) The autocorrelation plots corresponding to (a). (c) The relationship between the defocus distance and the pixel shift of the two copies.

Figure 2(a) shows the two-LED illuminated images at different focal planes (we only use the green channel for analysis). The corresponding autocorrelation plots are shown in Fig. 2(b). As the sample moves to the defocus positions, the captured images contain two copies of the sample separated by a certain distance. This two-copy separation can be directly recovered from the first-order peaks of the autocorrelation plot in Fig. 2(b). In Fig. 2(c), we show the calibrated relationship between the defocus distance of the sample and the separation distance between the two-copy. Figure 2(a) also demonstrates the long z-range of the reported approach. The depth of field of the employed objective lens is approximately 1.3 μm with the conventional Kohler illumination (i.e., turn on the big surface-mount white LED in Fig. 1). Thanks to the two partially coherent point LED sources, we can see that the out-of-focus contrast can be maintained over a long z range. This gives us the advantage of long focusing range compared to the existing method (Table 1).

A key innovation of the reported method is to set the stage to an offset defocus distance at the beginning. We set this offset distance to be 60 μm in our prototype setup. By setting this defocus distance, the sample position from -30 μm to +30 μm can be detected (i.e., the range from 30 μm to 90 μm in Fig. 2(c)). A larger offset distance results in a longer z-range for focal plane detection. On the other hand, a larger offset would reduce the focal plane detection accuracy. This is because the two LED sources are not ideal point sources and they have certain light emitting area. This point can be appreciated from Fig. 2(b2) and 2(b3). As we move the sample away from the focal position, the autocorrelation peaks reduce and the background increase in Fig. 2(b3).

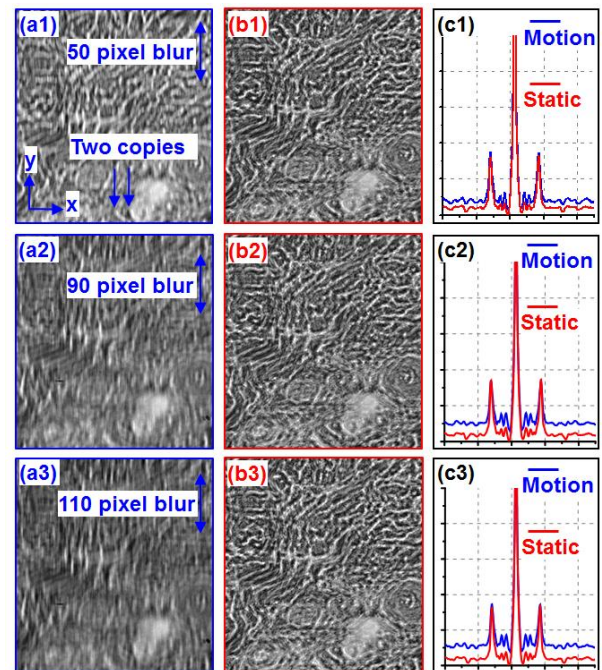

Fig. 3. (a) Images captured with continues motion in the y direction. (b) The corresponding static images. (c) The comparison of the autocorrelation curves between the motion-blurred and static images.

Another key innovation of the reported method is to enable focal plane detection with continues sample motion. This unique feature is based on the 1D autocorrelation curves in Fig. 2(b), where the captured images are in 2D and we only need to calculate the autocorrelation along the x direction. This allows us to introduce motion blur in the y direction for the captured images. Figure 3(a) shows the captured images with the sample in continuous motion along the y direction with different speeds. Figure 3(b) shows the corresponding static images. In Fig. 3(c), we compare the autocorrelation curves between the continuous-motion case and the static case. We can see that the reported method is robust against motion blur if the blur is along a direction perpendicular to the direction of the two-copy. The typical camera exposure time for the two LED point sources is 1 millisecond without setting readout gain. The 100-pixel motion blur allows us to move the sample at the speed of 20 mm/s without any gain setting. A higher speed can be easily achieved by reducing the exposure time with a readout gain.

Table 2. Summary of focusing errors

| Samples (1-10) | # of tiles | Static focusing error (µm) | 50-pixel motion blur (µm) | 90-pixel motion blur (µm) | 110-pixel motion blur (µm) |
|---|---|---|---|---|---|
| IHC slide (Cytokeratin) | 100 | 0.13 ± 0.10 | 0.17 ± 0.10 | 0.14 ± 0.08 | 0.13 ± 0.11 |
| H&E slide 1 | 50 | 0.07 ± 0.07 | 0.07 ± 0.06 | 0.13 ± 0.12 | 0.17 ± 0.12 |
| H&E slide 2 | 50 | 0.06 ± 0.05 | 0.07 ± 0.05 | 0.11 ± 0.06 | 0.18 ± 0.13 |
| H&E slide 3 | 50 | 0.06 ± 0.05 | 0.05 ± 0.05 | 0.23 ± 0.14 | 0.11 ± 0.12 |
| H&E slide 4 | 50 | 0.10 ± 0.08 | 0.10 ± 0.10 | 0.24 ± 0.10 | 0.17 ± 0.11 |
| H&E slide 5 | 50 | 0.07 ± 0.07 | 0.19 ± 0.08 | 0.17 ± 0.07 | 0.20 ± 0.13 |
| H&E slide 6 | 50 | 0.07 ± 0.06 | 0.07 ± 0.08 | 0.12 ± 0.10 | 0.11 ± 0.07 |
| H&E slide 7 | 50 | 0.04 ± 0.04 | 0.16 ± 0.05 | 0.08 ± 0.06 | 0.18 ± 0.08 |
| Human myocardial infarct sec | 50 | 0.10 ± 0.09 | 0.20 ± 0.07 | 0.14 ± 0.08 | 0.20 ± 0.14 |
| Unstained mouse kidney | 100 | 0.06 ± 0.06 | 0.06 ± 0.05 | 0.11 ± 0.07 | 0.21 ± 0.12 |
| Summary | 600 | 0.08 ± 0.07 | 0.11 ± 0.07 | 0.14 ± 0.09 | 0.17 ± 0.11 |

We have performed two experiments to quantify the focusing accuracy of the reported method. In the first experiment, we quantify the performance of the static mode, where the sample is not in continuous x-y motion while capturing images. The ground truth for the in-focus position is calculated based on an 11-point Brenner gradient method in an axial range of 5 µm (0.5 µm per step) [10]. The mean focusing error of the static mode is ~0.08 µm for 10 different pathology slides including a low-contrast immunohistochemistry (IHC) slide and an unstained mouse kidney section. The results are summarized in Table 2. In the second experiment, we quantify the performance of the continuous-motion mode. We adjust the LED intensity to achieve different motion blur cases in Table 2. The mean focusing error has been increased to ~0.17 µm, which is still much smaller than the depth of field. These two experiments have validated the accuracy of the reported method.

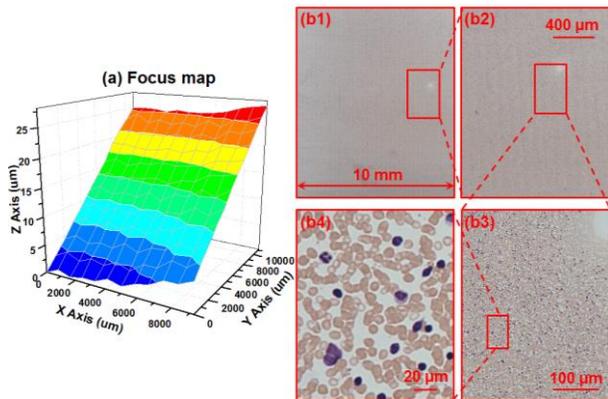

Fig. 4. (a) The generated focus map with continues sample motion. (b) The captured whole slide image using the focus map (also refer to http://www.gigapan.com/gigapans/200320)

As shown in Fig. 4(a), we create a focus map based on the reported method with continues sample motion (110-pixel motion blur). The corresponding high-resolution whole slide image is shown in Fig. 4(b), where all parts of the sample are in focus.

In summary, we have discussed a novel focus map surveying method for WSI with continuous sample motion. The innovation of the reported method is in twofold. First, we set a defocused offset distance to the stage before performing the focus map surveying. By doing so, we can generate out-of-focus contrast for transparent samples. This step also eliminates the use of additional cameras for focus point tracking. Second, we explore the unique 1D autocorrelation strategy of the reported method. By choosing the scanning direction to be perpendicular to the autocorrelation direction, we can minimize the effect of motion blur.

We envision several immediate applications of the reported method. First, we can use it for fluorescence WSI by simply employing two red (or near infrared) LED elements. In this case, the light from the two LEDs can pass through the filter cube for focus map surveying. No other modification is needed. Second, we can use it to correct the focus drift in time-lapse live cell experiments. The existing solution (such as Nikon Perfect Focus system) requires the user to choose an offset distance to a reference surface (for dry objectives, the reference surface is the air-dish interface). If the user wants to image many locations, the offset distance may vary because the thickness of the dish is not uniform. The reported method, on the other hand, is able to automatically pick the focal plane of the sample based on the coherent contrast. Third, we can implement it in a reflective mode. In this case, it may find wide applications in wafer and product inspection. Lastly, it can be implemented using linear sensor instead of 2D imager, and the autocorrelation can be implemented with an embedded system [11].

**Funding.** This work was in part supported by NSF 1555986, NSF 1700941, NIH R21EB022378, and NIH R03EB022144. G. Zheng has the conflict of interest with Clearbridge Biophotonics and Instant Imaging Tech, which did not support this work.